# Private Facial Diagnosis as an Edge Service for Parkinson's DBS Treatment Valuation

Richard Jiang, Paul Chazot, Danny Crookes, Ahmed Bouridane and M Emre Celebi

**Abstract** — Facial phenotyping has recently been successfully exploited for medical diagnosis as a novel way to diagnose a range of diseases, where facial biometrics has been revealed to have rich links to underlying genetic or medical causes. In this paper, taking Parkinson's Diseases (PD) as a case study, we proposed an Artificial-Intelligence-of-Things (AIoT) edge-oriented privacy-preserving facial diagnosis framework to analyze the treatment of Deep Brain Stimulation (DBS) on PD patients. In the proposed framework, a new edge-based information theoretically secure framework is proposed to implement private deep facial diagnosis as a service over a privacy-preserving AIoT-oriented multi-party communication scheme, where partial homomorphic encryption (PHE) is leveraged to enable privacy-preserving deep facial diagnosis directly on encrypted facial patterns. In our experiments with a collected facial dataset from PD patients, for the first time, we demonstrated that facial patterns could be used to valuate the improvement of PD patients undergoing DBS treatment. We further implemented a privacy-preserving deep facial diagnosis framework that can achieve the same accuracy as the non-encrypted one, showing the potential of our privacy-preserving facial diagnosis as an trustworthy edge service for grading the severity of PD in patients.

—————————— ◆ ——————————

## 1 INTRODUCTION

FACIAL diagnosis dates back to 2000 years ago in Traditional Chinese Medicine(TCM) practice [1-2]. Recently, scientists have shown that facial phenotyping [3-8] can be used to diagnose over 200 rare genetic diseases, reviving this ancient technology as a new method potentially for diagnosing of a wide range of diseases. In addition to applying to rare diseases, facial diagnosis can be related to many common illnesses, such as the Parkinson's disease (PD) [7-14]. It has been widely reported that PD has an apparent impact on the faces of PD patients, like many other neurodegenerative diseases [11, 12].

PD can affect natural facial expressions in addition to gross motor skills. This phenomenon, called "facial-masking" (hypomimia), is a common sign of early PD, arising prior to major motor and non-motor symptoms. This is due to a reduction of automatic and controlled expressive movement of facial musculature, creating an appearance of apathy, social disengagement or compromised cognitive status. Such facial symptoms can have significant negative effects on mental health, as it distorts the perceived emotional responses of the individual, which can lead to misinterpretation by friends, family and the public. Hence, via facial diagnosis, a robust computational model can capture subtle variations of facial features, establish quantifiable evidence and measure/predict the developmental stages of Parkinson's Diseases.

- *Richard Jiang is with LIRA Center, Lancaster University, Lancaster, LA1 4YW, United Kingdom. Correspondence E-mail: r.jiang2@lancaster.ac.uk.*
- *Paul Chazot is with School of Biosciences, Durham University, Durham, DH1 3DE, United Kingdom*
- *Danny Crookes is an emeritus professor with Department of Computer Science, Queen's University Belfast, Belfast, BT9 1NN, United Kingdom.*
- *Ahmed Bouridane is with Dept. of Computer and Information Sciences, Northumbria University, Newcastle, NE1 8ST, United Kingdom.*
- *M Emre Celebi is with Dept. of Computer Science, University of Central Arkansas, Conway, AR 72035, USA.*

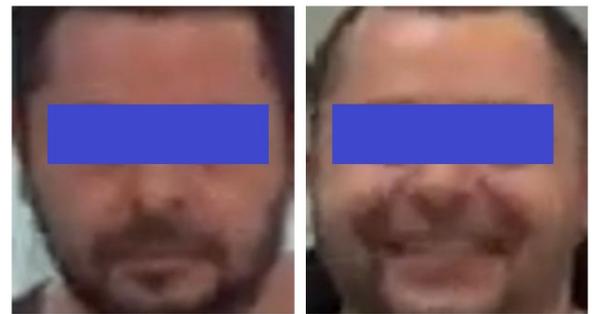

Figure 1. Facial images of a PD patient before and after Deep Brain Stimulation treatment.

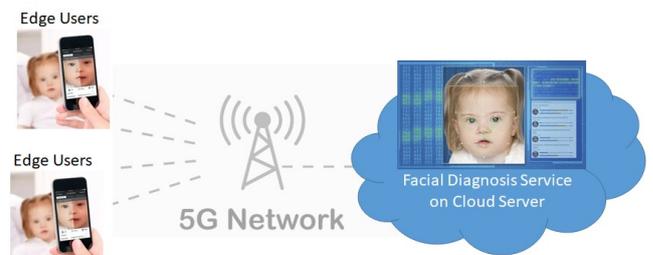

Figure 2. Facial diagnosis over 5G AIoT edge service

Parkinson's Diseases [9-14] is a heterogeneous disease (no case being exactly the same), with motor symptoms including tremor, rigidity, bradykinesia, dyskinesia (drug-induced), and non-motor symptoms, including dementia, anxiety, depression, sleep dysfunction, psychosis, persistent pain, delirium, gambling compulsion etc. At the early developmental stage, these symptoms are subtle and occasional. Via automated facial diagnosis, individuals in a range of settings can be monitored and diagnosed quickly and scored on the risk of potential phenotypes of PDs for doctors/caregivers to consider social interventions and palliative therapies. Facial diagnosis [1-8] can provide a cost-effective staff-free automated diagnosis platform via 24/7 real-time online surveillance of patients in care with no waiting time. Potentially, it will enable early diagnosis for preventive treatments over a large population in care and help provide accurate phenotyping for timely precision medicine or treatment. To bring facial diagnosis into the PD analysis, there are two challenges the researchers and doctors need to overcome:

1) First, we need proof-of-concept experiments to show that facial images can be practically used to classify PD patients with suitable accuracy.
2) Secondly, due to privacy issues, the facial biometrics of patients need to be restricted to private use instead of circulating over Internet servers, and a privacy-preserved framework is needed for automated facial diagnosis services over cloud-based networks.

In this paper, taking the PD treatment via Deep Brain Stimulation (DBS) as our case study, we aim to develop a private IoT-based facial diagnosis framework to evaluate the improvement of DBS treatment on patients by comparing facial biometrics before and after DBS.

Fig.1 shows the sample facial images of a patient before and after DBS treatment, respectively. Facial images are deformable, and key features need to be subtly discriminated. Cross-subject classification becomes more challenging particularly, when the training dataset consists of totally different subjects from the test dataset. Although there are reports that links facial features to PDs [11, 12], up to now it has not been certain if we can use facial images to grade the severity of PD patients via machine learning. In this paper, we will prove this concept using a deep learning based framework.

The scenario of cloud-based automated facial diagnosis can be depicted in Fig.2. A face photo of the patient can be collected at an edge-based Artificial-Intelligence-of-Things (AIoT) device such as a mobile phone, and the photo is then sent to the server of the service provider. Due to privacy concerns and legistation, patients (or hospitals) may not allow private facial biometric data to be uploaded to the server located in an external business. A privacy-aware framework is then required for the purpose of commercialized facial diagnosis services to offer a widely accessible platform for point-of-care (POC) AIoT end users.

In our proposed solution, we exploit Partial Homomorphic Encryption (PHE) to develp a light-weighted privacy-preserving deep learning based framework for facial diagnosis. The reasons for choosing PHE instead of Full Homomorphic Encryption (FHE) include: 1) PHE has been applied for medical applications for years [15~19] and is relatively mature with much lower costs; 2) It is less compute-intensive than full homomorphic encryption (FHE) [19-21] and thus easier to deploy on AIoT devices; 3) PHE based machine learning has recently been successfully utilized on facial images [22-25]. Hence, a PHE based privacy-preserving deep learning framework is a rational choice for our facial diagnosis application.

Our contributions in this work include three key aspects, as detailed below:

1) For first time, we demonstrated that machine learning can effectively discriminate between facial images of a patient before and after DBS treatment.
2) We implemented a novel lightweight AIoT-based framework for privacy-preserving deep learning based facial diagnosis, which is secure from an information theory standpoint;
3) We also demonstrate that our proposed framework can achieve the equivalent accuracy in comparison with the original non-privacy-preserving facial diagnosis algorithm.

We implemented our initial demo with a PHE library and validated our implementation as the proof-of-concept on a video datasets of 52 PD patients who received the DBS treatments.

The remainder of the paper is organized as follows. Section 2 reviews the existing relevant work. Section 3 gives background on PHE and describes its implementation. Section 4 presents the proposed AIoT-based privacy-preserving deep learning framework. Section 5 shows the experimental results. Finally, Section 6 concludes the paper.

## 2 REVIEW OF PRIVACY-PRESERVING LEARNING

To protect the priacy of shared data over the Internet, it has been widely researched to use various encryption methods [15-28] to secure the data under various hostile or curious conditions. While privacy can be mostly assured by a sophisticated encryption scheme, the data services that need to carry out various computations on the data cannot work due to the encryption. To solve this issue, homomorphic encryption schemes have become a popular way to protect the private data while enabling computation, such as machine learning, to operate on encrypted data.

Partial Homomorphic Encryption, as its name suggests, can maintain only some homomorphic features in its encryption, while full homomorphic encryption is expected to be able to carry out all equivalation arithmetic operations in its encrypted domain. Typical PHE algorithms include Pallier, ELGamel, and RSA [21]. Well known FHE libraries include SEAL, HElib and TFHE [20]. Table I gives a comparison of several existing open encryption algorithms. We can see that division and comparison are not yet available in either PHE or FHE algorithms. Unlike FHE algorithms, PHE algorithms cannot support bitwise operation, exponentiation, negation, and add/subtract with plain.

Although PHE does not provide all operations to implement deep learning in the encrypted domain, in this paper, we will demonstrate a new scheme to fulfill the privacy-preserving deep learning without resorting to expensive FHE ones. By taking the Pallier cryptosystem as our encryption method, we will show a full implementation of privacy-preseving deep neural networks using PHE based

**Table I Comparison of FHE and PHE algorithms.**

| Ops. | FHE | | | PHE | | |
|---|---|---|---|---|---|---|
| | SEAL | HElib | TFHE | Pallier | ELGamal | RSA |
| Add/Sub | √ | √ | √ | √ | × | × |
| Multiply | √ | √ | √ | × | √ | √ |
| Division | × | × | × | × | × | × |
| Compare | × | × | × | × | × | × |
| Matrx Ops | √ | √ | × | × | × | × |
| Exp | √ | √ | × | × | × | × |
| Square | √ | √ | √ | × | √ | √ |
| Neg | √ | √ | × | × | × | × |

on an AIoT framework.

There are a number of reported applications of privacy-preserving machine learning [15]-[28]. The majority of these are developed for distributed setting where different parties hold parts of the training database and securely train a common classifier without each party needing to disclose its own training data to other parties. Recently, Pallier encryption has been applied successfully for facial classification using classic linear machine learning methods [22-25] such as SVM, LDA, etc. In our work, we aim to utilze deep learning for our facial diagnosis service, because deep learning has achieved greater robustness and better accuracy in pattern classification [3, 6-8].

There have been reports of various FHE based deep learning implementations [20, 29-33]. However, it is yet challenging to implement the nonlinear functions in neurons since FHE is fundamentally based on linearization lattice. As shown in Table I, current FHE methods cannot yet cope with divide and compare [20]. Its implementation of exponention is based on the repeating squaring algorithm, which causes extra computational complexity and accuracy concerns, and can be applied only to positive number exponentiation [20]. Due to the complexity, most reported implementations assumes the linear approximation of nonlinear functions, with no guarantee of the control of extra errors [29-33]. On the other side, PHE is relatively less compute-intensive [22-24]. Hence, in our application to PD, we choose PHE to implement a lightweight framework for privacy-preserving deep learning.

While the current trend is to provide AI-enabled medical diagnosis as a service [15, 16, 18, 19, 21], the needs to preserve the privacy of the data comes with data sharing over distributed systems, cloud-based services, 5G/6G communication, Internet-of-Things, etc.. There are many reported frameworks to try to meet the various needs. In [27], the encrypted data from clients are decrypted on the server side before feeding into the deep neural networks and the final results are encrypted before sending back to clients. However, this implies the server can know the client's private data with no privacy protection, though it may be secure to third parties in the communication nodes. In [31], the encryption is performed partially on the layers of deep neural networks. However, asymmetric encryption, although data is hidden from each other, lacks robustness due to the exposure of public keys [34-35], leaving it vulnerable if sufficient attempts are made. Hence, these proposed schemes over multi-party communication will not be secure in the information theoretical sense.

In this work, we are concerned with facial diagnosis over AIoT services. When patients or hospitals send their data over to the commercial service provider, they may not want to risk compromising the privacy of their patients. On the other side, the service provider wants to keep their learned model safe, and does not want to be exposed to an end user. Hence an information theoretically secure scheme is required for such automated medical data services.

Encrypted facial recognition has been implemented with many algorithms [22-25]. Privacy-preserving data classification algorithms suitable for a client-server model were studied in [36-38]. The client-server model substantially reduces the computational and communication overhead at the client since it needs to interact with only one server compared to a more distributed setting. Outsourcing the clinical decision support system to a third-party server without violating the client's privacy was studied in [15-19, 21-24, 27-33]. In this paper, in contrast to these previous works, we propose a lightweight privacy-preserving medial diagnosis protocol based on encrypted deep learning over an information theoretically secured AIoT framework. We will demonstrate that our proposed facial diagnosis framework achieves the desired privacy requirements without degrading the classification performance, and the computational and communication complexity are under control.

## 3 PRIVACY-PRESERVING DNN IMPLEMENTATION

Considering the scenario that AIoT users send their encrypted facial images to the automated medical diagnosis server hosted by a commercial service provider, we need to find a robust privacy-preserving scheme to guarantee data privacy in the system. To achieve this, as detailed in the above sections, we choose the Paillier cryptosystem as our encryption engine, which is partially homomorphic.

### 3.1 The Pallier Cryptosystem

Typical PHE algorithms [21-24] include Unpadded RSA, ElGamal encryption, Goldwasser–Micali cryptosystem, Benaloh cryptosystem, Paillier cryptosystem, etc.. The Paillier cryptosystem [24] is an additive homomorphic encryption method. The Paillier cryptosystem, invented by and named after Pascal Paillier in 1999, is a probabilistic asymmetric algorithm for public key cryptography. The problem of computing $n^{th}$ residue classes is known to be computationally difficult. The decisional composite residuosity assumption is the intractability hypothesis upon which this cryptosystem is based.

Like most other RSA-like algorithms, the Paillier encryption needs a pair of keys, namely public and private keys. The keys are generated via the process below:

1) Public keys: Choose two large prime numbers $p$ and $q$ randomly and independently of each other such that: $gcd(pq, (p-1)(q-1)) = 1$. Compute $n=pq$ and $\lambda=$lcm$(p-1, q-1)$.

2) Select a random integer $g \in Z_{n^2}^*$, and calculate the modular multiplicative inverse:

$$\mu = \mod\left(\frac{n}{\mod(g^\lambda, n^2) - 1}, n\right)$$

(1)

Here the division gives the integral quotient.

With the above two steps, the generated public keys are $\{n, g\}$ and the private keys are $\{\lambda, \mu\}$.

After keys are produced, the encryption step will be decided by public keys only. Assuming $t$ is the message to encrypted and satisfies $0 \le t < n$, and $r$ is a random number and satisfies $0 \le r < n$ and $r \in Z_n^*$, the cipher text can be obtained as:

$$c = [\![t|r]\!] = \mod(g^t r^n, n^2), \text{ and } c \in Z_{n^2}^*$$

(2)

Here, $[\![*]\!]$ denotes the encryption. The random number $r$ will make the Pailier encryption as non-deterministic.

Every time with the same keys, the same text $t$ will have different cipher text $c$.

Inversely, the decipher step will be based on the private keys, and

$$t = \langle c \rangle = \mod\left(\frac{\mod(c^\lambda, n^2) - 1}{n} \mu, n\right)$$
(3)

Here, $\langle * \rangle$ denotes decryption in this paper. As it can be seen pointed out by Paillier in 1999 [21], decryption here is "essentially one exponentiation modulo $n^2$".

### 3.2 Homomorphic Properties

A notable feature of the Paillier cryptosystem is its homomorphic properties along with its non-deterministic encryption. The Pallier encryption function is additively homomorphic, and the product of two ciphertexts will decrypt to the sum of their corresponding plaintexts,

$$[\![ t_1 + t_2 ]\!] = \mod([\![ t_1 | r_1 ]\!] \cdot [\![ t_2 | r_2 ]\!], n^2)$$
(4)

With the above equation, we can easily obtain the sum of two encrypted numbers without deciphering them.

Besides the additive homomorphic property, the encrypted numbers in the Paillier cryptosystem can implement the negation homomorphic property as,

$$[\![ t_1 \times t_2 ]\!] = \mod([\![ t_1 ]\!]^{t_2}, n^2)$$
(5)

Here, it is noted that $t_2$ must be in its original value. Given the Paillier encryptions of two messages there is no known way to compute an encryption of the product of these messages without knowing the private key.

Based on the above partial multiplicative homomorphic property, the encrypted numbers in the Pallier cryptosystem can implement a form of the negation homomorphic property,

$$[\![ -t_1 ]\!] = \mod([\![ t_1 ]\!]^{-1}, n^2)$$
(6)

This is particularly useful since in artificial neurons we have negative weights and bias.

Sometimes we may need to add the ciphertext with a plain text, which could be done simply by,

$$[\![ t_1 ]\!] + t_2 = [\![ t_1 ]\!] + \mod([\![ 1 ]\!]^{-t_2}, n^2)$$
(7)

Here, a plaintext is encoded by taking advantage of $[\![ 1 ]\!]$ and Eq.(5), without knowing the encryption keys (includeing public keys).

### 3.3 Encrypted Faces

Consider a pattern or image consisting of $W \times H$ pixels, where each pixel is represented by its intensity value $x_i$. Using Eq.[2], we can then obtain the encrypted big integer $z_i = [\![ x_i ]\!]$ for each pixel. Normalizing $\{z_i\}$ by $n^2$, we can obtain the whole encrypted images shown in Fig.3. The original images are on the left, and the corresponding encrypted patterns are visualized on the right.

As we can observe from Fig.3, the encrypted patterns are no longer visually identifiable, though each pattern contains exactly the same amount of information as the original images. However, without knowing the keys, no one can decipher the patterns. Hence, data security and privacy can be safely guaranteed, unlike other alternative arrangements in many private learning methods [15-19] that may compromise the security over shared keys between multiple parties.

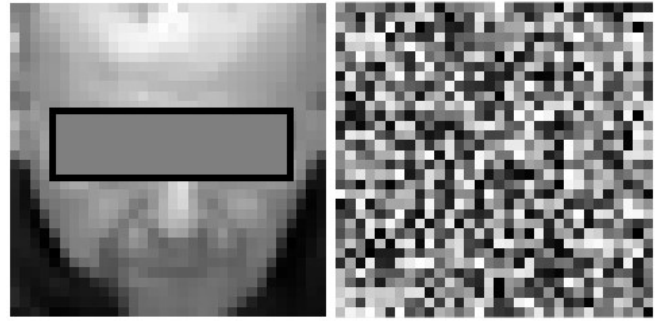

Figure 3. Encrypted digital patterns

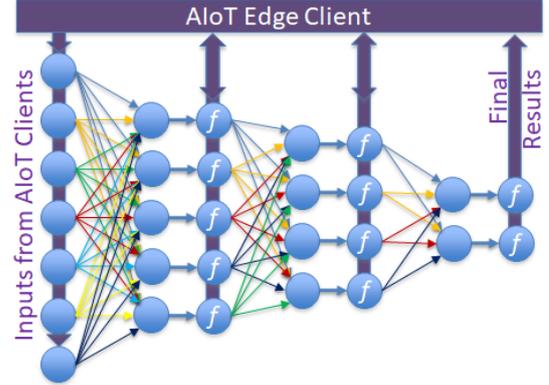

Figure 4. An encryption based private deep neural network running in a server coomunicating with AIoT edge clients.

### 3.4 Encrypted Neurons without Keys

Artificial neurons are a simplified model of biological neurons that contain multiple synaptic inputs and one axon output. The model of an artificial neuron consists of a weighted-sum process that can be descrbed as,

$$y_j = \sum_{i=1}^{K} w_i x_i + b_j$$
(8)

where $x_i$ are the input signals and $y_j$ is the summation result that will be capped by an activation process, typically the sigmoid function,

$$z_j = \frac{1}{1 - e^{-y_j}}$$
(9)

$z_j$ denotes the final output from the $j$-th neuron in a layer.

If we have encrypted inputs $[\![ x_i ]\!]$, then the addition process in the above neuron model will be described as,

$$\begin{aligned}
[\![ y_j ]\!] &= \left[\!\!\left[ \sum_{i=1}^{K} w_i x_i + b_j \right]\!\!\right] \\
&= \mod\left(\left[\!\!\left[ \sum_{i=1}^{K} w_i x_i \right]\!\!\right] \cdot [\![ b_j ]\!], n^2\right) \\
&= \mod\left(\left(\prod_{i=1}^{K} [\![ w_i x_i ]\!]\right) \cdot [\![ b_j ]\!], n^2\right) \\
&= \mod\left(\left(\prod_{i=1}^{K} [\![ x_i ]\!]^{w_i}\right) \cdot [\![ b_j ]\!], n^2\right)
\end{aligned}$$
(10)

Hence, the calculation can be easily carried out directy on the encrypted inputs. Here, we may need to encrypt the bias value in the learned neuron model. To make it simple, we can take advantage of Eq.(5) and obtain,

$$\llbracket b_j \rrbracket = \llbracket 1 \times b_j \rrbracket = \text{mod}(\llbracket 1 \rrbracket^{b_j}, n^2) \qquad (11)$$

Hence, we do not need any encryption keys to fulfil the abve calculation if the big integer value of $\llbracket 1 \rrbracket$ is given.

The calculation of the activation function in Eq.(9) on encrypted data is a great challenge because it needs to carry out divisions and exponentiations in the encrypted domain. Some attempts adpoting various linear approximation have been reported [32, 33]. However, these approximations may introduce unexpected errors. Similarly, the implementation of softmax functions on encrypted data suffers from similar difficulties.

In this paper, we provide an AIoT-based solution to this challenge in privacy-preserving deep learning. Fig.4 shows its schematic view. While the learned model is stored in the server in a layer-wise structure, the activation functions are computed via queries to the AIoT clients in a distributed computing framework, and the computational load is shared by the server and the clients. More details can be found in the following section.

## 4 PROPOSED DEEP FACIAL DIAGNOSIS FOR AIoT

### 4.1 Overall Framework

The overall framework of our proposed AIoT-based facial diagnosis is detailed in the diagram in Fig.5. The central medical service with well-trained deep neural networks resides on a cloud-based Honest-But-Curious (HBC) server, and AIoT edge clients can upload their encrypted data to the server. The encrypted data is then processed in a layer-wise mode through deep neural networks, and each layer is followed by the activation through the communication with the AIoT clients, as shown in the architecture in Fig.4. The server may host many requests from different AIoT clients in corresponding sessions, as far as it can handle by its computing capability.

The detailed algorithms for the AIoT clients and the facial diagnosis server are listed in List I. In an AIoT server, the Pallier cryptosystem is set up to encrypt the query data for privacy protection. The encrypted data is then sent to the server side for medical diagnosis. The server will then carry out the computation in a layer-wise mode, and at each layer, will send the tentative outputs from each layer back to the client for the computation of activation or softmax, which needs both division and exponential operations. Once the computation at the server side reaches the last layer of deep neural networks, thefinal result will be sent back to the client and the communication session will be terminated.

In this framework, the server can maintain multiple sessions with different clients, as far as multi-threading is allowed on the cloud server. It is worth noting that all keys (both public and private) are kept only on the client itself, and the computation at the server side is keyless and only needs an encpted big integer $\llbracket 1 \rrbracket$.

### 4.2 Privacy-Preserving Secure AIoT Edge Service

As shown in Fig.5, the proposed AIoT medical diagnosis framework is based on two-party communication in its computation. Though a server can host many clients, the communication session is based on a client with a communication channel to the server. Hence, there is a need to look into the data privacy issue of the two-party secure

### List I. Algorithms of the proposed framework

**Algorithm 1** AIoT Client
**Inputs:**
   (1)  Sample pair $D_1$ and $D_2$ for diagnosis
   (2)  Client ID *uid* and password *pwd*
   (3)  Server queries with *cmd* and *Xl*
**Procedure:**
1:  setup bitlength *Bl* and and float point length *Fl*
2:  setup communication with server using *uid* and *pwd*
3:  **loop** over $D_1$ & $D_2$:
4:    create public & private keys of Pallier cryptosystem
5:    encrypt $D_k$ and 1, send to the diagnosis server
6:    **wait** for the returned data from server: (*cmd, Xl*)
7:    **if** *cmd* is "sigmoid"
8:      calculate $sigm(\langle Xl \rangle)$, encrypt & send to server
9:    **if** *cmd* is "softmax"
10:     calculate $softm(\langle Xl \rangle)$, encrypt & send to server
11:    **if** *cmd* is "results"
12:     decrypt $Xl$, $R_k = \langle Xl \rangle$;
13: **end loop**
14: **Compare** the scores $R_1$ and $R_2$:
15:  **if** $R_1 > R_2$: $D_1$ is before and $D_2$ is after DBS;
15:  **else:** $D_2$ is before and $D_1$ is after DBS;
16: **close the communication & return**

**Algorithm 2** Keyless Private Facial Diagnosis
**Inputs:**
   (1)  Client lists with IDs and Passwords
   (2)  Pretrained weights $w_{ij}$ and bias $b_j$
   (3)  From client: Query data $\llbracket D_k \rrbracket$ and $\llbracket 1 \rrbracket$
**Procedure:**
1:  Setup DNNs with weights $w_{ij}$ and bias $b_j$;
2:  **wait** for the client request:
3:    verify user and setup the session;
4:    **wait** for the client data $\llbracket D_k \rrbracket$ and $\llbracket 1 \rrbracket$:
5:    **loop** over layers of the diagnosis DNN:
6:      calculate $\llbracket y_j \rrbracket$ in layer-wise;
7:      shuffle $\llbracket y_j \rrbracket$ over $j$ randomly;
8:      calculate $\llbracket z_j \rrbracket$ by querying the client (*sigmoid*);
9:    **end loop**
10:   calculate $\llbracket r_j \rrbracket$ by querying the client (*softmax*);
11:   send the final results back to the client
12:   **close** the session
13: **return** to step 2

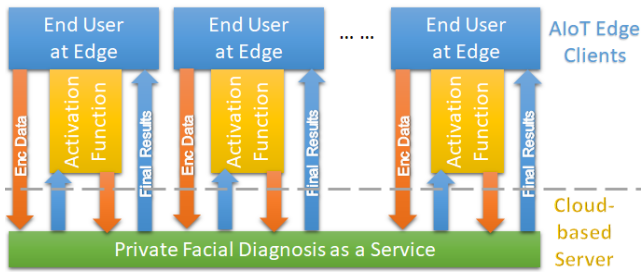

Figure 5. Schematic view of the proposed privacy-protected client-server framework for AIoT edge services.

computation.

### 4.2.1 Privacy with Information Theoretical Security

An algorithm is defined as information-theoretically secure if its security is proved to match with the derived concept from information theory. The information-theoretical security in communication was termed by Shannon to prove that the one-time pad system achieves perfect security subject to the following two conditions [39]:

1) the key which randomizes the data should be random and should be used only once;
2) the key length should be at least as long as the length of the data.

If any algorithm randomizes its parameter and satisfies the above conditions, its parameters cannot be unmasked by an adversary even when the adversary has unlimited computing power. In this case, the system is guaranteed to have information theoretical security. If for example both the plain text and ciphertext have 1024 bits, the prior probability (probability for a particular ciphertext out of $2^{1024}$ possible ciphertexts) and posterior probability (probability of inferring/mapping a ciphertext in a random domain to a plaintext domain) are equal, and there is no advantage for an adversary to get higher posterior probability than prior probability.

In the following sections, we will analyze whether our algorithm in List I is vulnerable to any privacy leakage. Our algorithm is based on two party computation over the AIoT scheme. As defined by Goldreich [40], privacy for secure two-party computation will be achieved if a secure two-party protocol cannot reveal more information to a semi-honest party than the information that can be inferred by looking at that party's input and output. To verify whether the proposed AIoT computation satisfies the privacy definition, we will show the inputs and outputs to and from the client and server, respectively, in the algorithm in List I, and identify what is already known to the client and the server. Following this, if we can prove that nothing else can be inferred other than the known input and output with higher posterior probability than prior probability, the proposed algorithm in List I then satisfies with the privacy definition based on information theory.

The ultimate aim for the client is to keep the test image and the classification result away from the server while the server wants to keep the classification model parameters away from the client. In the communicaion, the client sends only the encrypted messages to the server with no public keys. From these inputs, the server knows only the size of the test image and the length of encrypted data. This is not a privacy leakage since the server knows the size of the images when training the classifier.

### 4.2.2 Privacy-Preserving Secure Computation at Edge

The overall framework needs its computation to be based on secure operations in both AIoT clients and the server. An honest-but-curious server/client follows the protocol and takes no actions beyond those of an honest server/client. In the proposed framework, let us assume the client knows a vector (data) **a** and sends its ciphertext $[\![\mathbf{a}]\!]$ to the server. The whole system will expect the ciphertext is unbreakable in both the server and any third party. Since the encrypted DNNs at the server side need neither private keys nor public keys, the security of the client data in the proposed AIoT-oriented keyless medical diagnosis system is therefore guaranteed, though asymmetric encryption can be breakable if public keys are available [34, 35].

**Theorem 1**. The proposed two-party AIoT protocol in List I is information-theoretically secure i.e., the server cannot infer the plaintexts in the client's input vector $[\![\mathbf{a}]\!]$ and the client will only learn the final result but not the model parameters in the server.

**Proof**: To validate the security we consider both the client and server are honest-but-curious (HBC) i.e., they will follow the procedures but try to learn about each other's inputs, intermediate values and results. Let us show that the algorithm in List I is information-theoretically secure for the following two cases.

### 4.2.3 Honest-But-Curious AIoT-oriented Server

The server computes the results in their encrypted form. Each layer performs a weighted summation, as shown in Eq.(8). The HBC client sends its encrypted data to the server, and the server computes the weighted sum outputs and shuffle them via one time pad, and send them back to the client for the computation. Because the inputs and the outputs are shuffled each time using different one-time pads (step 7 in algorithm 2, List I), the client cannot accumulate the knowledge to solve $w_{ij}$ aand $b_j$ from Eq.(8). Hence, there is no privacy leak on model parameters such as weights and bias values in layers, and the communication from the server to the client is perfectly information theoretical secure in term of the server's private model.

### 4.2.4 Honest-But-Curious AIoT Edge Clients

The main purpose in the proposed framework is to protect the data privacy of the individual client from the public medical diagnosis server. In List I, a client encrypts its data and sends the data to the server, and the server carries out its computation on encrypted data only, and return the encrypted result back to the client, which implies that only the client can know the final results via decryption.

However, as has been suspected [34, 35], asymmetric encryption itself may not be information theoretical secure, since the factorization of a large prime number can be solved via some algorithms [34] in limited time, if the public keys can be known. However, in our proposed algorithms in List I, neither private keys nor public keys are shared with the server. The curious server cannot infer the keys simply from the ciphertext.

It is worth noting that in the proposed scheme, the encryption of 1 needs to be sent to the server. Considering $t=1$, we have

$$c1 = [\![1|r]\!] = \mathrm{mod}(gr^n, n^2), \text{ and } c \in Z_{n^2}^* \qquad (12)$$

and,

$$1 = \langle c1 \rangle = \mathrm{mod}\left(\frac{\mathrm{mod}(c^\lambda, n^2) - 1}{n}\mu, n\right) \qquad (13)$$

In the above equations, $g$, $r$, $n$, $c$ and $\lambda$ are all unknown to the server side, and these parameters are changed every time based on one time pad strategy. Hence, the server can not accumulate more knowledge about the encryption parameters from the next ciphertext, and the private data from the client will be information theoretical secure to the server side.

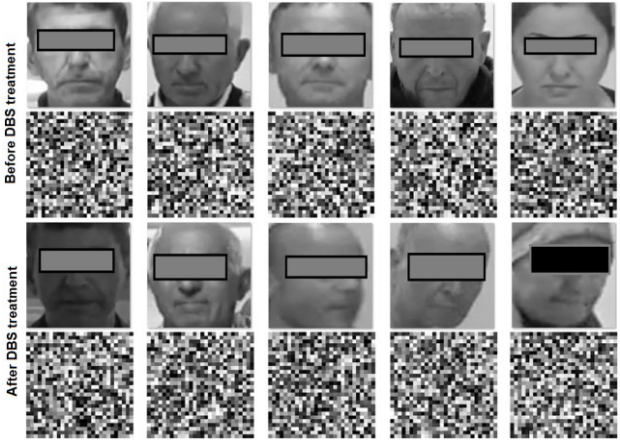

Figure 6. PD DBS facial dataset.

# 5 EXPERIMENTS

## 5.1 PD DBS Dataset

In our experiment, we use a video dataset consisting of 52 PD patients. Two video shots were taken per patient before and after DBS treatment. From videos, we extract 2345 facial images for our tests, as shown in Fig.6. The facial images were cropped from video shots and resized to 32×32 pixels. The dataset is then split into two subsets – training dataset and test dataset, each having 24 PD patients.

In our experiments, it is assumed that the training dataset is owned by the server to train the deep neural networks for medical diagnosis, and the test datatset is owned as private data by AIoT clients. The experiments were set up as a cross-subject facial diagnosis test, while training faces and test faces are from different subjects/patients. Such cross-subject issues are often considered more challenging than the same subject tests (such as recognize a face of a person already in the training dataset).

Our experiments are set up as below. Given two images of a patient, one from before the DBS treatment and one from after the treatment, we want to know:

1) Can facial features be used to clearly discriminate which one is before or after DBS treatment?
2) Can the proposed lightweight PHE-based solution work properly?
3) Can our PHE-based solution achieve an accuracy similar to its original non-privacy-preserving peer?

To answer the above questions, our experiments are detailed in the following sections.

## 5.2 Facial Diagnosis using Deep Autoencoder

Although facial expression is known to be associated with the degree of Parkinson's, it is yet not used on the evaluation of DBS treatment. To achieve this purpose, we aim to exploit deep neural networks [41] to carry out the facial diagnosis on the patients.

In our experiments, our targets is to demonstrate the proposed PHE based lightweight implementation of deep neural networks over encrypted data. We choose Deep Autoencoder as our classifier because it is the first reported

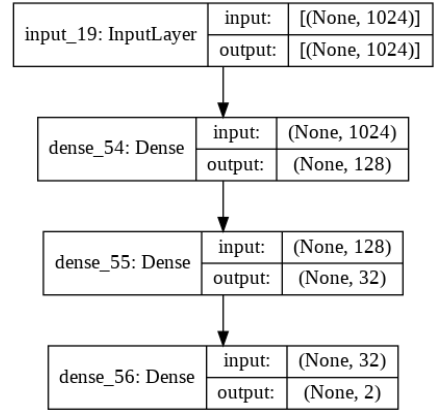

a) The structure of the deep encoder as a classifier

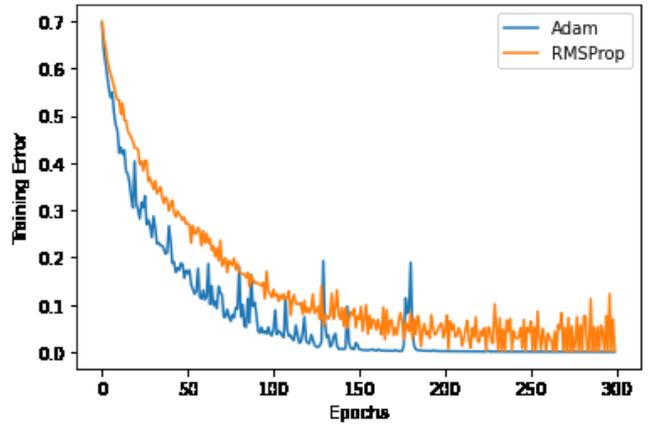

b) Training the classifer in the server side. Two traiing methods, Adam amd RMSProp were used.

| Method | Precision | Recall | Acc | F1 |
|--------|-----------|--------|-----|-----|
| Adam | 94.88% | 94.28% | 94.71% | 94.54% |
| RMSProp | 96.20% | 95.00% | 95.22% | 95.10% |

c) Evaluation on the cross-subject dataset

**Confusion Matrix**

|  | Target Class 1 | Target Class 2 |  |
|--|--|--|--|
| **Output Class 1** | **651** 55.6% | **32** 2.7% | 95.3% 4.7% |
| **Output Class 2** | **24** 2.0% | **464** 39.6% | 95.1% 4.9% |
|  | 96.4% 3.6% | 93.5% 6.5% | **95.2%** **4.8%** |

d) Confusion matrix on the test dataset of the RMSProp Model

Figure 7. Train the deep encoder classifies on the server side.

method that enlightened the research on deep learning [41]. Here, we only need the encoder part, as shown in Fig.7-a, a four layer deep neural network architecture with two hidden layers. It takes all dimensions of the data as its inputs, and then reduces the number of neurons via the propagation from layer to layer, from 1024, 128 to 32. At the output layer with a softmax activation, it has two neurons standing for two classes: before DBS treatment (with apparent PD symtoms) and after DBS treatment (with little PD symtoms).

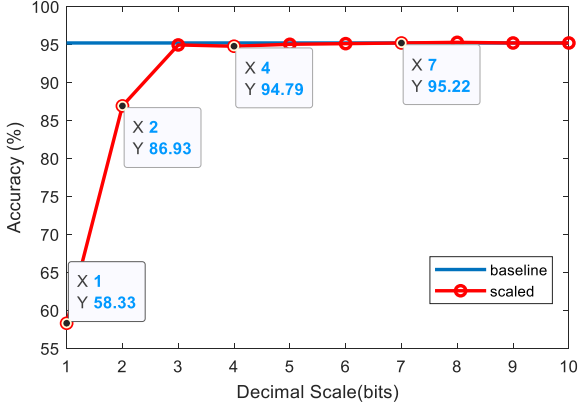

Figure 8. Test accuracy over the decimal precision of the encrypted DNN model parameters.

**Table II. Computational time versus the bit length on server and client, measured in seconds averaged per test image in our experiment.**

| Bits | 256 | 512 | 768 | 1024 |
|---|---|---|---|---|
| Server (s) | 42 | 64 | 101 | 146 |
| Clients (s) | 1 | 2 | 5 | 9 |
| Total (s) | 43 | 66 | 106 | 155 |

**Table III The estimated computational complexity**

| Layers | Server | Client |
|---|---|---|
| Input | 0 | $m_0(2C_e + C_m)$ |
| Layer 1 | $m_1 m_0 (C_e + C_m)$ | $3m_1(C_e + C_m)$ |
| Layer 2 | $m_2 m_1 (C_e + C_m)$ | $3m_2(C_e + C_m)$ |
| Layer 3 | $m_3 m_2 (C_e + C_m)$ | $3m_3(C_e + C_m)$ |
| Total | $(m_3 m_2 + m_2 m_1 + m_1 m_0)$ $\times (C_e + C_m)$ | $(2m_0 + 3m_1 + 3m_2 + 3m_3)C_e$ $+ (m_0 + 3m_1 + 3m_2 + 3m_3)C_m$ |

**Table IV The communication complexity**

| Layers | S2C | C2S |
|---|---|---|
| Input layer | 0 | $m_0 C_{Big}$ |
| Layer 1 | $m_1 C_{Big}$ | $m_1 C_{Big}$ |
| Layer 2 | $m_2 C_{Big}$ | $m_2 C_{Big}$ |
| Layer 3 | $m_3 C_{Big}$ | 0 |
| Total | $(m_1 + m_2 + m_3) \times C_{Big}$ | $(m_0 + m_1 + m_2) \times C_{Big}$ |

In our implementation, the input layer has 32×32 = 1024 nurons, the 2nd layer has 128 neurons with the sigmoid activation, and the 3rd layer has 32 neurons with the ReLU activation. The 4th layer is the softmax layer with two neurons to outpt the likelihoods of before and after DBS treatment.

We used the training dataset (facial images of 26 patients) trained a four layer autoencoder, and obtained the model parameters. Fig.7-b) shows the convergence of loss rates in the training process. Here, two training methods were applied, Adam and RMSProp. We can see the model trained by RMSProp achieved better accuracy.

Fig.7-c) shows the valuation on two trained models in term of accuracy, recall, precision and F1 score based on the output likelihoods on two classes. We can see the RMSProp based model achieved a bit better results consistently on all measures. Hence, we use the RMSProp model as our facial diagnosis service model. Fig.7-d shows its confusion matrix of the test results. We can clearly see that the model achieved a nice accuracy over 95%, implying that there are sufficiently explicit facial features to discriminate a PD patient form a recovered patient.

It is worth to note that the diagnosis is based on a single image instead of long shot videos, and the test subjects are all new to the trained model. From the test results, we can clearly see the trained model can explicitly discriminate the faces of patients before and after DBS treatment, suggesting its potential as an automated medical diagnosis tool for evaluating the severity of PD patients.

### 5.3 Encrypted Facial Diagnosis over AIoT Edge

Following the test in section 5.2, we implemented the encrypted counterpart of the above facial diagnosis model and compare the test results on the encrypted test dataset. Here, a concern issue is about the number of decimal digits that may scale down the model parameters (weights, bias, etc.). The encrypted data from a client was fed to the encrypted DNN model on the server side, and we want to examine if the algorithms in List I can achieve similar accuracy as the original non-privacy-preserving model even with a reduced or limited number of decimal digits.

Fig.8 shows the comparison of the results of the encrypted implementation with different decimal precisions. The decimal precision varies from 1 bit to 10 bit, and we valuated the classification accuracy on the encrypted data. We can see that we decimal bits is reduced to 1bit ($2^{-1}$ or 0.5), the accuracy is only 58.33%. When the decimal bits were increased to 7 bits, the encryption-based model achieved the same accuracy (95.22%) as its non-privacy preserving original model. Compared with the length og big integer, 7 bits for decimal digits could be a small cost for the overall encryption scheme.

Table II also gives the experimental results on the computational time in the client and the server, respectively, over various bit lengths of the big integer. Here both client and server were running on a GPU-powered laptop with 3.6GHz CPU. The results show that in the same session, most computational work load (~95%) was assigned to the server side, which is a benefit for AIoT end users offered by the service provider. Typically, the computational time can be linked to the computational complexity, as analyzed below.

### 5.4 Computational Complexity

The computational complexity of the algorithm in List I can be estimated as the costs of two types of big integer operations: multiplication operations and modulo exponentiation as $C_m$ and $C_e$, respectively. Table III list out the estimated complexity of the proposed schemes for both the server (S) and client (C) with regards to the number of neurons per layer ($m_0$, $m_1$, etc.) and the number of synaptic inputs to each neuron (namely the number of neurons in the previous layer, in the architecture illustrated in Fig.4).

In our implementation, the input layer has 32×32 neurons, layer 1 has 128 neurons, layer 2 has 32 neurons and the final layer has 2 neurons. Hence, the total computational costs can be estimated as:

Server side: (1024×128+128×32+32×2)➔135232×($C_e$+$C_m$)

Client side: 2534×$C_e$+1510×$C_m$

We can see most computational work load is assigned to the server side. Assuming $C_e \approx C_m$, the ratio of the work load between the server and a client is 135232/4044 ≈ 36.1, which implies the server copes with 97% of the total computation. This theoretical analysis matches well with the experimental results (~95%) in Table III. Since the cloud-based server has much better computing resources, such a client-server split is reasonable for an AIoT edge-oriented framework.

## 5.5 Communication Complexity

We can measure the total communication complexity in terms of data being communicated between the server and client. In our algorithm, the transferred data between client and server are all in the form of big integer, such as 1024-bit format, as detailed in the secure two-party algorithm proposed in List I. The estimated communication load is summarized in Table IV based on a layer-wise estimation in term of the communication costs of a single 1024-bit big integer, $C_{Big}$.

Similarly, since the input layer has 32×32 neurons, layer 1 has 128 neurons, layer 2 has 32 neurons and the final layer has 2 neurons, the total communication costs can be estimated as:

Server to client: 162×$C_{Big}$

Client to server: 1184×$C_{Big}$

In total, the communication bandwidth needs to cope with 1346×$C_{Big}$ = 1346×1024bits = 1.378Mbits. If we process video streams at 25 frames per seconds, the lowest bandwidth needs to be aroun 35 Mbps. Currently 5G-based AIoT can easily achieve 100+ Mbps. Hence, the communication complexity in our proposed AIoT framework is achievable even in the case of considering real-time video streaming, although our facial diagnosis carries out on single image based classification.

## 6 CONCLUSION AND FUTURE WORK

In conclusion, we have demonstrated a novel AIoT-oriented medical diagnosis framework using deep learning and PHE that can achieve Shannon's information theoreitical security. In our proposed scheme, the encrypted DNNs at the server side were implemented without the needs for encryption keys, guaranting the client data has information theoretical security; the intermediate results from the server to a client were shuffled, making the model unknown to the client, too. Hence, the security at both server and client ends is guaranteed.

It is also worth noting that, with the 95.22% accuracy of our facial diagnosis model, we verified that facial images can serve as an important cue to evaluate the severity of PD patients, showing that it has potential for a new facial diagnosis tool in medical applications.

Moreover, we demonstrated that PHE methods such as the Paillier cryptosystem can be used effectively to implement deep learning with encrypted data, and with its information theoretical security we provide a ground for the easy implementation of privacy-preserving deep neural networks, particularly for the needs of medical services over the Internet.

It is worth to highlight that our cross-subject tests implied there are exlicit discriminative facial features that are linked to the severity of PDs. One of our future tasks is to explicitly expand our deep facial model to an explainable model, to help identify which part of facial features are linked to the severity of PDs.

In this work, we implemented our facial service as a privacy-preserving model. However, the training on the server side was carried out on the unencrypted data which were assumed to be owned by the service provider. A future task is to expand this training process (such as Adam and RMSProp methods) on encrypted data and allow privacy-preserving model training. It can be valuable to extend to private training data and novel frameworks such as federated learning, while data providers may not allow the service provider fully own those private data.

## ACKNOWLEDGMENT

This work was supported in part by Grant (EP/P009727/1) from the EPSRC and Grant (RF-2019-492) from the Leverhulme Trust, and the Grant 1946391 by the US National Science Foundation.

## REFERENCES

[1] F. Cheung, "TCM: Made in China," Nature, vol. 480, no. 7378, pp. 82–83, Dec. 2011.

[2] X. Li, *et al*, "Face Parsing for Traditional Chinese Medicine Inspection via a Hybrid Neural Network," in IEEE Access, 2020.

[3] Gurovich, Y., *et al*. Identifying facial phenotypes of genetic disorders using deep learning. *Nat Med* 25, 60–64 (2019).

[4] S. Richmond, et al, "Facial Genetics: A Brief Overview", Front. Genet., Oct. 2018.

[5] Daniel J. M. Crouch, Bruce Winney, Joseph Kittler, *et al*, "Genetics of the human face: Identification of large-effect single gene variants", PNAS, 2018.

[6] Delude CM (2015) Deep phenotyping: The details of disease. Nature 527, S14-S15.

[7] G Storey, R Jiang, et al, "3DPalsyNet: A Facial Palsy Grading and Motion Recognition Framework using Fully 3D Convolutional Neural Networks", 2019, IEEE Access.

[8] G Storey, A Bouridane, R Jiang, "Integrated Deep Model for Face Detection and Landmark Localisation from 'in the wild' Images", 2018, IEEE Access.

[9] Dorsey, E. Ray et al. 'Deep Phenotyping of Parkinson's Disease'. 1 Jan. 2020 : 1 – 19.

[10] Krüger R, *et al*. "Classification of advanced stages of Parkinson's disease: translation into stratified treatments", *J Neural Transm (Vienna)*. 2017.

[11] Argaud S, et al., "Facial emotion recognition in Parkinson's disease: A review and new hypotheses", *Mov Disord*. 2018. doi:10.1002/mds.27305

[12] Dawn Bowers, et al, "Startling facts about emotion in Parkinson's disease: blunted reactivity to aversive stimuli", Brain, Dec. 2006.

[13] Klein C, Westenberger A. Genetics of Parkinson's disease. *Cold*

*Spring Harb Perspect Med*. 2012;2(1).

[14] Rial D, et al. "Behavioral Phenotyping of Parkin-Deficient Mice: Looking for Early Preclinical Features of Parkinson's Disease". PLoS ONE, 2014.

[15] X. Liu, R. Lu, J. Ma, L. Chen and B. Qin, "Privacy-Preserving Patient-Centric Clinical Decision Support System on Naïve Bayesian Classification," in IEEE Journal of Biomedical and Health Informatics, vol. 20, no. 2, pp. 655-668, March 2016, doi: 10.1109/JBHI.2015.2407157.

[16] X. Yang, R. Lu, J. Shao, X. Tang and H. Yang, "An Efficient and Privacy-Preserving Disease Risk Prediction Scheme for E-Healthcare," in IEEE Internet of Things Journal, vol. 6, no. 2, 3284-3297, April 2019, doi: 10.1109/JIOT.2018.2882224.

[17] X. Liu, R. H. Deng, K. R. Choo and Y. Yang, "Privacy-Preserving Outsourced Support Vector Machine Design for Secure Drug Discovery," in IEEE Transactions on Cloud Computing, vol. 8, no. 2, pp. 610-622, 1 April-June 2020, doi: 10.1109/TCC.2018.2799219.

[18] J. Feng, L. T. Yang, Q. Zhu and K. R. Choo, "Privacy-Preserving Tensor Decomposition Over Encrypted Data in a Federated Cloud Environment," in IEEE Transactions on Dependable and Secure Computing, vol. 17, no. 4, pp. 857-868, 1 July-Aug. 2020, doi: 10.1109/TDSC.2018.2881452.

[19] H. Zhu, X. Liu, R. Lu and H. Li, "Efficient and Privacy-Preserving Online Medical Prediagnosis Framework Using Nonlinear SVM," in IEEE Journal of Biomedical and Health Informatics, vol. 21, no. 3, pp. 838-850, May 2017, doi: 10.1109/JBHI.2016.2548248.

[20] X. Sun, *et al*, "Private Machine Learning Classification Based on Fully Homomorphic Encryption," IEEE Trans Emerging Topics in Computing, vol. 8, no. 2, 2020.

[21] Y. Rahulamathavan, S. Veluru, R. Phan, J. Chambers, and M. Rajarajan, "Privacy-preserving clinical decision support system using Gaussian kernel based classification," IEEE Journal of Biomedical and Health Informatics, vol. 18, no. 1, pp. 56–66, Jan. 2014.

[22] Z. Erkin, M. Franz, J. Guajardo, S. Katzenbeisser, I. Lagendijk, and T. Toft, "Privacy-preserving face recognition," in Proc. 9th Int. Symp. Privacy Enhancing Technol., 2009, pp.235-253.

[23] Y. Rahulamathavan, R. Phan, J. Chambers, and D. Parish, "Facial expression recognition in the encrypted domain based on local fisher discriminant analysis," IEEE Trans. Affective Computing, vol. 4, no. 1, pp. 83–92, Jan.-Mar. 2012.

[24] Rahulamathavan, Y., Rajarajan, M. "Efficient Privacy-preserving Facial Expression Classification," IEEE Trans. Dependable and Secure Computing, Vol.14, No.3, May 2017, 326-338.

[25] R. Jiang, et al, "Privacy-Protected Facial Biometric Verification Using Fuzzy Forest Learning," IEEE Transactions on Fuzzy Systems, 1 Aug. 2016.

[26] H. Lipmaa, S. Laur, and T. Mielikainen, "Cryptographically private support vector machines," in Proc. 12th ACM SIGKDD Int. Conf. Knowl. Discovery Data Mining, Aug. 2006, pp. 618-624.

[27] M. Barni, T. Bianchi, D. Catalano, M. Di Raimondo, R. D. Labati, P. Failla, D. Fiore, R. Lazzeretti, V. Piuri, F. Scotti, and A. Piva, "Privacy-preserving fingercode authentication," in Proc. 12th ACM Workshop Multimedia Security, 2010, pp. 231–240.

[28] W. Du and Z. Zhan, "Building decision tree classifier on private data," in Proc. IEEE Int. Conf. Privacy, Security Data Mining-Volume 14, 2002, pp. 1-8.

[29] Hesamifard, E., Takabi, H. & Ghasemi, M. CryptoDL: deep neural networks over encrypted data. Preprint at https://arxiv.org/abs/1711.05189 (2017).

[30] Dowlin, N. et al. CryptoNets: applying neural networks to encrypted data with high throughput and accuracy. In Proc. 33rd Int. Conf. Machine Learning Vol. 48 201–210 (PMLR, 2016).

[31] Qiao Zhang, Cong Wang, HongyiWu, Chunsheng Xin and Tran V. Phuong, "GELU-Net: A Globally Encrypted, Locally Unencrypted Deep Neural Network for Privacy-Preserved Learning", Proceedings of the Twenty-Seventh International Joint Conference on Artificial Intelligence (IJCAI-18).

[32] Karthik Nandakumar，Nalini Ratha, Sharath Pankanti，Shai Halevi，"Towards Deep Neural Network Training on Encrypted Data", Proceedings of the IEEE/CVF Conference on Computer Vision and Pattern Recognition (CVPR) Workshops, 2019

[33] Théo Ryffel1, Edouard Dufour-Sans, Romain Gay, Francis Bach and David Pointcheval, "Partially Encrypted Machine Learning using Functional Encryption", 33rd Conference on Neural Information Processing Systems (NeurIPS 2019), Vancouver, Canada.

[34] Shor, P.W. (1994). "Algorithms for quantum computation: discrete logarithms and factoring". Proceedings 35th Annual Symposium on Foundations of Computer Science. IEEE Comput. Soc. Press: 124–134.

[35] Amico, Mirko; Saleem, Zain H.; Kumph, Muir (2019-07-08). "An Experimental Study of Shor's Factoring Algorithm on IBM Q". Physical Review A. 100 (1): 012305.

[36] Zhao, C. et al. Secure multi-party computation: theory, practice and applications, Inform. Sci. 476, 357–372 (2019).

[37] Evans, D., Kolesnikov, V. & Rosulek, M. A Pragmatic Introduction to Secure Multi-Party Computation (NOW, 2018).

[38] Jagadeesh, K. A., Wu, D. J., Birgmeier, J. A., Boneh, D. & Bejerano, G. Deriving genomic diagnoses without revealing patient genomes. Science 357, 692–695 (2017).

[39] C. E. Shannon, "Communication theory of secrecy systems," Bell Syst. Tech. J., vol. 28, no. 4, pp. 656-715, 1949.

[40] O. Goldreich. (1998, Sept.) Secure multiparty computation [Online]. Available: http://www.wisdom.wei zmann.ac.il/oded/pp.html

[41] LeCun Y, Bengio Y, Hinton G. Deep learning. Nature 2015;521(7553):436-444.

**First A. Author** All biographies should be limited to one paragraph consisting of the following: sequentially ordered list of degrees, including years achieved; sequentially ordered places of employ concluding with current employment; association with any official journals or conferences; major professional and/or academic achievements, i.e., best paper awards, research grants, etc.; any publication information (number of papers and titles of books published); current research interests; association with any professional associations. Author membership information, e.g., is a member of the IEEE and the IEEE Computer Society, if applicable, is noted at the end of the biography.

**Second B. Author Jr.** biography appears here.

**Third C. Author** biography appears here.